\pdfoutput=1

\documentclass[11pt]{article}

\usepackage{acl}

\usepackage{subfigure}
\usepackage{times}
\usepackage{latexsym}
\usepackage{CJKutf8}
\usepackage[T1]{fontenc}

\usepackage[utf8]{inputenc}

\usepackage{microtype}

\usepackage{amsmath}

\usepackage{amsfonts}

\usepackage{soul}
\usepackage{tikz}
\usepackage{longtable}
\usepackage[export]{adjustbox}

\usepackage{makecell}

\usepackage{cleveref}
\usepackage{xcolor, color}
\usepackage{graphicx}
\usepackage{epstopdf}
\usepackage{multirow}
\usepackage{booktabs}
\usepackage{comment}
\usepackage{float}
\usepackage{enumitem}
\usepackage{ctable} 
\usepackage{anyfontsize} 
\usepackage{titlesec}
\titlespacing*{\paragraph}{0pt}{0.5ex plus 0ex minus 0ex}{1em}

\title{Analyzing the Intensity of Complaints on Social Media}

\author{
Ming Fang$^1$\quad Shi Zong$^1$\quad Jing Li$^{2}$\quad Xinyu Dai$^1$\quad Shujian Huang$^1$\quad Jiajun Chen$^1$\\
$^1$National Key Laboratory for Novel Software Technology, Nanjing University, China\\
\texttt{fangming@smail.nju.edu.cn,} \\
\texttt{\{szong, daixinyu, huangsj, chenjj\}@nju.edu.cn}\\
$^2$Department of Computing, The Hong Kong Polytechnic University, HKSAR, China\\
\texttt{jing-amelia.li@polyu.edu.hk} \\
  }

\usepackage[backgroundcolor=white]{todonotes}

\begin{document}

\begin{CJK}{UTF8}{gbsn}

\maketitle

\begin{abstract}
Complaining is a speech act that expresses a negative inconsistency between reality and human expectations. While prior studies mostly focus on identifying the existence or the type of complaints, in this work, we present the first study in computational linguistics of measuring the \textit{intensity} of complaints from text. Analyzing complaints from such perspective is particularly useful, as complaints of certain degrees may cause severe consequences for companies or organizations.
We create the first Chinese dataset containing 3,103 posts about complaints from Weibo, a popular Chinese social media platform. These posts are then annotated with complaints intensity scores using Best-Worst Scaling (BWS) method.
We show that complaints intensity can be accurately estimated by computational models with the best mean square error achieving 0.11.
Furthermore, we conduct a comprehensive linguistic analysis around complaints, including the connections between complaints and sentiment, and a cross-lingual comparison for complaints expressions used by Chinese and English speakers.
We finally show that our complaints intensity scores can be incorporated for better estimating the popularity of posts on social media.\footnote{Our annotated corpus is publicly available at \url{https://github.com/nlpfang/complaint_intensity}.}
\end{abstract}

\section{Introduction}

Complaining is caused by the gap between reality and people's expectations \cite{olshtain1985complaints}. \citet{brown1987politeness} state that the purpose of complaining is not to confirm that the two parties have reached an agreement but to face-threatening acts. People use complaints to express their concerns or dissatisfaction based on the severity and urgency of situations. 

Researchers from linguistics and psychology have long pointed out that people may shape their complaints to varying degrees \citep{olshtain1985complaints, jenkins1979levels, trosborg2011interlanguage}. \citet{leech2016principles} classifies complaints as conflicting speech acts. Mild complaints can reach the purpose of venting emotions to promote mental health, but serious complaints can lead to hatred and even bullying behaviors \cite{iyiola2013relationship}.
In computational linguistics, prior studies primarily focus on building automatic classification models for identifying the existence of complaints \cite{preotiuc2019automatically}. Most recently, \citet{jin-aletras-2021-modeling} provided a dataset annotated with different severity levels of complaints based on the theory of pragmatics, including four distinct categories ``no explicit reproach'', ``disapproval'', ``accusation'' and ``blame''. 

Among these studies, we note one missing piece is to measure the \textit{intensity} of complaints. 
To illustrate this point, consider two sentences from the newest annotated dataset from \citet{jin-aletras-2021-modeling}: \emph{can i complain to you about the coffee i just received ?} and \emph{virgin media as usual full of lies lies lies ! ! !}. Although these two complaints may have the same type ``accusation'', it is clear that they are different regarding the degree of complaints. 
As another example, \textit{totally not cool.} and \textit{please reply my dm asap ! ! !} are both classified as ``disapproval''. However, the latter makes a stronger complaint.
Analyzing different complaints levels can also be beneficial. Companies need to regularly monitor the feedback from users, as certain complaints may significantly impact the reputation of their products. Organizations or governments need to monitor people's biggest complaints to understand their urgent needs.

In this work, we analyze the intensity of complaints on social media. To the best of our knowledge, it is the first computational linguistics study that tries to automatically capture the complaints intensity from text.
We present the first Chinese complaints intensity dataset, consisting of 3,103 posts from Weibo. 
We then show that the complaints intensity can be measured from text by building computational models.
We further demonstrate the necessity and importance of understanding complaints intensity. 
This includes a detailed analysis that distinguishes the differences between our complaint intensity scores and original sentiment scores.
As a pilot study for complaints in Chinese, we also perform a cross-lingual analysis to understand the differences in the complaint expressions used in Chinese and English. We have some interesting empirical findings. For example, we observe that English speakers tend to use more ironic expressions than Chinese speakers. 
Finally, we show how our annotated corpus can help predict the popularity of posts on social media.

\section{Data}
\label{sec:data}

In this section, we present the first Chinese dataset that is annotated towards the intensity of the complaints reflected from text. 

\subsection{Data Collection}
\label{sec:collection}

We collect data from Weibo,\footnote{\url{www.weibo.com}} a famous social media platform in China that is similar to Twitter. As posts about complaints only account for a minority of the total posts on Weibo, in this work we consider education domain -- an area that is the primary focus for most families in China, which generally raises hot debates and complaints about current education policies.
We selected a set of keywords related to complaints, including {\small 抱怨 (\emph{complaint})}, {\small 不公平 (\emph{unfair})}, and {\small 举报 (\emph{report})}.
We then randomly sampled out 5 hashtags around these keywords and collected Weibo posts from these hashtags. We collected a total of 4,490 Weibo posts from August 2020 to May 2021. 

\paragraph{Pre-processing.} 
We notice that the hashtag on Weibo is usually a sentence (in ``\#...\#'' format), rather than a phrase like its Twitter counterparts. To ensure a certain amount of content generated by users, we filtered out posts with less than 10 words and more than 200 words (without hashtags). For each post, we removed the name of the author, location tags, and URLs. We also converted emoticon into text format. Finally, 3,103 Weibo posts remain for annotation. \Cref{tb:breakdown_stats} shows the breakdown statistics in our corpus.

\begin{table}[h!]
\small
\centering
\resizebox{0.495\textwidth}{!}{
\begin{tabular}{p{0.48\textwidth}c}
\toprule
\textbf{Hashtag}     & \textbf{Num.} \\\midrule
\#代表建议让学生在校内完成家庭作业\#\\ (\emph{\#The representative suggested that students should complete their homework on campus\#}) & 762 \\\hline
\#江苏明确教师不得用手机布置作业\#\\ (\emph{\#Jiangsu Province makes it clear that teachers are not allowed to use mobile phones to assign homework\#})  & 534 \\\hline
\#院士不建议普通孩子学奥数\#\\(\emph{\#Academician does not recommend ordinary children to learn Mathematical Olympiad\#})   & 627 \\\hline
\#西安外国语大学封闭管理\#\\
(\emph{\#Close management of Xi'an International Studies University\#})  & 598 \\\hline
\#人大法硕复试30余人成绩0分\# \\
(\emph{\#More than 30 people scored 0 in the postgraduate examination of law at Renmin University\#})   & 582 \\\midrule
\textbf{Total} & 3,103 \\
\bottomrule
\end{tabular}
}
\caption{Hashtags and number of collected Weibo posts in our annotated corpus.}
\label{tb:breakdown_stats}
\end{table}

\subsection{Data Annotation}

\paragraph{Complaints Levels.}

Our goal is to measure the intensity of complaints from text. We adopt the definition from \citet{jenkins1979levels}, which quantifies the complaints into five levels, as shown in \Cref{tab:level_complaint}. Higher levels indicate stronger complaints.

\begin{table}[h!]
\small
\centering
\begin{tabular}{c|l}
\toprule
\textbf{Level} & \textbf{Description} \\\midrule
1 & a little anxiety and disgust \\
2 & deliberately expressing anxiety \\
3 & actively looking for ways to solve anxiety\\
4 & frustrated behavior \\
5 & depression, fear, and despair\\\bottomrule
\end{tabular}
\caption{Guideline used in annotation process for distinguishing different levels of complaints, adopted from \citet{jenkins1979levels}.} 
\label{tab:level_complaint}
\end{table}

In pilot studies, we test the feasibility of using these levels as the annotation guideline for the annotators, along with the potential mismatches between Chinese and English speakers. We observe that annotators are able to make comparison between complaints of different degrees. As discussed later, our annotations also achieve high agreement between annotators. 

\paragraph{Best-Worst Scaling (BWS).}

In this work, we annotate the complaint intensity using Best-Worst Scaling, proposed by \citet{louviere1991best}.
We choose this method as it can produce more stable and fined-grained scores than directly scoring \cite{kiritchenko-mohammad-2017-best}.
We note similar methods have been applied to various tasks, including measuring offensiveness \cite{hada2021ruddit} and intimacy \cite{pei-jurgens-2020-quantifying} in the computational linguistic literature. 

In BWS annotation, annotators are provided with 4-tuples randomly generated that meet certain criteria.\footnote{Requirements are: (1) no two 4-tuples are the same; (2) no two posts within a 4-tuple are identical; (3) each post appears approximately in the same number of 4-tuples; (4) each pair of posts appears approximately in the same number of 4-tuples.}
Annotators are then asked to select the strongest complaint item and the weakest complaint item within each 4-tuple. 
In practice, we randomly generated 2$n$ distinct 4-tuples, with $n$ being the number of posts. This amount of tuples is considered to be sufficient for getting reliable scores from annotation \cite{kiritchenko-mohammad-2017-best}.
We assign the complaint intensity score for each post by using the percentage of strongest cases minus the weakest cases, ranging from -1 to 1.

\begin{table*}[h!]
\centering
\small
\resizebox{0.99\textwidth}{!}{
\begin{tabular}{cp{0.9\textwidth}c}
\toprule
\textbf{Bin} & \textbf{Weibo posts}  & \textbf{Scores}\\\midrule
\multirow{2}{*}{1} & {\small 以前在没有手机的年代，孩子们都是自己记作业。我觉得有助于形成自我管理能力 (\emph{In the era when there were no mobile phones, children kept their homework by themselves. I think it helps to form self-management ability.})}     &  \multirow{3}{*}{-1}                                              \\\midrule
\multirow{2}{*}{2}   & {\small 最好的解决办法就是没有家庭作业，对老师对家长都好 (\emph{The best solution is to have no homework, which is good for teachers and parents.}} & \multirow{2}{*}{-0.56}   \\\midrule
3   & {\small 现在的学生作业为啥都得用手机做? (\emph{Why do students have to use mobile phones to do their homework now?})}           & +0.12       \\\midrule
4 & {\small 我是真的觉得用手机交作业很烦 (\emph{I find it really annoying to hand in homework with my mobile phone.})}      & +0.4  \\\midrule
\multirow{2}{*}{5}   & {\small 气死了！食堂涨价，超市关门，就没人管理嘛? (\emph{Mad! The price of the canteen increases, and the supermarket closes, no one manages it?})} & \multirow{2}{*}{+1}    \\
\bottomrule
\end{tabular}
}
\caption{Sample posts with complaints intensity scores. We divide our scoring scale (from -1 to 1) into 5 bins of size 0.4 (i.e., bin 1 refers to scores ranging from -1.0 to -0.6, bin 2 from -0.6 to -0.2, etc.). More examples are provided in \Cref{tab:sample_more} in \Cref{subsec:keywords}.}
\label{tab:sample_tweets}
\end{table*}

\paragraph{Annotation Quality.}

To ensure the quality of our annotations, we manually annotated 100 posts and asked all annotators to annotate them beforehand. We removed annotators whose accuracy is less than 70\% on these golden annotations. To get highly reliable results, we got each tuple annotated by 3 annotators. In total, we received more than 14,000 annotations from 15 annotators.

We follow the literature \cite{kiritchenko-mohammad-2017-best} and measure the quality of annotations by using score-to-half reliability (SHR). SHR score is calculated by randomly splitting all the tuples into two halves and then computing the correlation between these two groups. We repeat the above process 100 times. The average SHR score is 0.91, which indicates strong reliability.

\subsection{Data Analysis}

\begin{figure}[h!]
    \centering
    \includegraphics[width=0.48\textwidth]{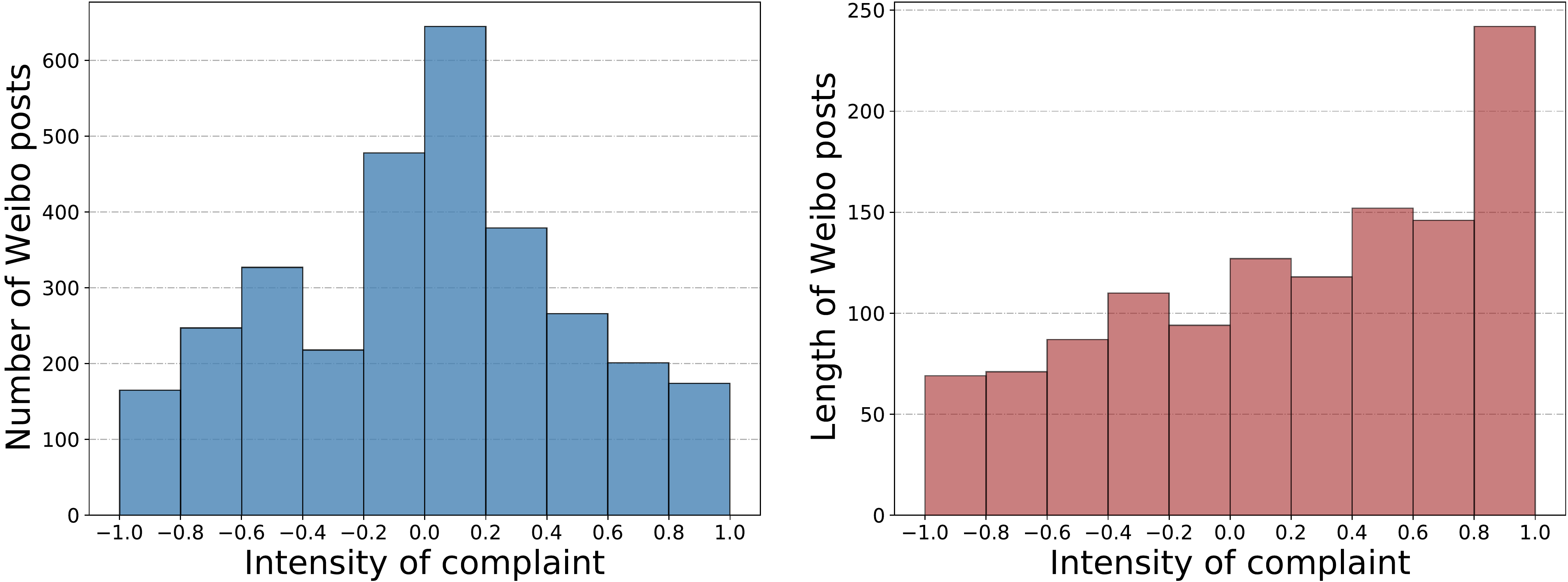}
    \caption{Distribution for the number and length of posts over complaints intensity in our corpus.}
    \label{fig:distribution}
\end{figure}

We first analyze the distribution for the annotated complaint intensity scores in our corpus. As shown in \Cref{fig:distribution} (Left), we observe a normal distribution for the number of posts across different complaint scores, with most of the posts having intensity of complaints within -0.2 and 0.2. 

We also observe that the length of complaint posts (intensity$>$0) is longer than that of non-compliant posts (intensity$<$0) in \Cref{fig:distribution} (Right). By examining our data, we observe it is because stronger complaints contain more details with more aspects. 
For example, in bin 5 of \Cref{tab:sample_more} (in \Cref{subsec:keywords}), the target of complaint changes from {\small 学校 (\emph{school})} to dissatisfaction with {\small 图书馆 (\emph{library})} and even accuses the behavior of {\small 门卫 (\emph{security guard})}. This is the halo effect in psychology: if something leaves a wrong impression, everything related to it becomes terrible. On the contrary, we observe that most non-complaining posts contain only plain expressions, and people will not describe too much after expressing their opinions on the matter.

\section{Predicting the Intensity of Complaints}

In \Cref{sec:data}, we have a dataset annotated with the intensity of complaints from -1 to 1. We now build computational models for predicting the intensity of complaints of a given post.

\subsection{Models}

\paragraph{Support Vector Regression (SVR).} 
We use support vector regression as our first baseline model. We experiment with two different input sentence representations: bag of \{2,3,4\}-gram features and 300-dimensional GloVe embeddings \cite{pennington2014glove}. Results in \Cref{tab:main_result} use an RBF (Radial Basis Function) kernel. In practice we observe similar results using other kernels (e.g., linear kernel).

\paragraph{Bidirectional LSTM.} We also experiment with a bidirectional Long Short-Term-Memory (Bi LSTM) model. The LSTM and the average pooling layer concatenation are passed through a linear layer with a tanh activation, producing a score between -1 and 1. We use two sets of embedding for input layers: Glove \cite{pennington2014glove} and BERT for embedding \cite{li2020sentence}. The attention mechanism is also considered. Other hyperparameters for the models are a batch size of 64, a learning rate of 1e-3, 13 epochs with early stopping, and dropout of 0.5 to avoid overfitting. 

\paragraph{Pre-trained Models.} We finally experiment with pre-trained models, including BERT \cite{devlin2018bert}, RoBERTa \cite{liu2019roberta}, and ERNIE \cite{zhang-etal-2019-ernie}. For all pretrained models, we add a linear layer as a regression layer to the model.
We then fine-tune these models using a mean square error loss objective. We set batch size to be 16 and learning rate to be 2e-5. The model is trained for 3 epochs. All hyperparameters are selected using a held-out dev set.

\subsection{Experiments}
\label{sec:experiment}

We evaluate the model performances for our complaint intensity prediction task in two settings: (1) mix hashtag, where we combine Weibo posts from different hashtags together, and (2) cross hashtag, where the posts for train, dev and test sets are separately from different hashtags. We use Pearson's correlation and MSE (Mean Square Error) as metrics for all our experiments.

\paragraph{Mix Hashtag.}
We combine Weibo posts from different hashtags together and then split them into 80\% for training, 10\% for validation, and 10\% for test. Results are shown in \Cref{tab:main_result}. We observe that RoBERTa outperforms all other models and reaches Pearson up to 0.79, followed by the LSTM model. The SVR model has the worst performance. 

\paragraph{Cross Hashtag.}
We choose four of total five collected hashtags as the train and development set, and hold out the rest one for test. We report the average value after five experimental runs in \Cref{tab:main_result}. We observe under cross hashtag setting, models get comparable performances to mix hashtag setting. It indicates models seem to learn common linguistic cues between different hashtags.

\begin{table}[h!]
\centering
\resizebox{0.48\textwidth}{!}{
\begin{tabular}{l|cccc}
\toprule
\multirow{2}{*}{\textbf{Models}} &
  \multicolumn{2}{c}{\textbf{Mix Hashtag}} &

  \multicolumn{2}{c}{\textbf{Cross Hashtag}} \\
& $r$ & MSE &  $r$ & MSE \\\midrule
SVR (\{2,3,4\}-gram) & 0.36   & 0.46      & 0.35   & 0.46 \\
SVR (GloVe)        & 0.49   & 0.36     & 0.47  & 0.38     \\
\midrule
LSTM (GloVe)        & 0.69   & 0.24      & 0.65   & 0.27 \\
LSTM Attn (Glove)     & 0.72    & 0.22     & 0.70   & 0.25\\
LSTM (BERTembed)    & 0.76   & 0.15     & 0.75   & 0.16 \\
\midrule
ERNIE               & 0.76   & 0.14     & 0.76   & 0.14 \\
BERT                & 0.77   & 0.20     & 0.75   & 0.23 \\
RoBERTa             & 0.79  & 0.11     & 0.78   & 0.11  \\   
\bottomrule
\end{tabular}
}
\caption{Pearson's $r$ and Mean Square Error (MSE) on two datasets for predicting the intensity of complaints.}
\label{tab:main_result}
\end{table}

\subsection{Error Analysis}
\label{sec:error_analysis}

We perform an error analysis to shed light on the limitations of our best-performing model. A prediction is defined to be wrong when the difference between ground truth and the predicted score is greater than 0.5. 
We randomly sample 100 errors and manually inspect them. 
All errors are divided into three categories: 43\% of errors are because of irony expression in complaints, 29\% are due to implicit expressions, and 28\% are due to the insufficient and vague expressions.

\paragraph{Ironic Expressions.} 
We observed that most errors happen when the posts contain ironic expressions. Users use positive words such as ``perfect'' or ``great'' to express dissatisfaction, which are misleading models to ignore the implicit complaints. A typical example is as follows: {\small 食堂的涨价消息比封校政策来的快, 这学校真好 (\textit{The news of price increase in the cafeteria is coming faster than the school closure policy. This school's management is really good})}.\footnote{During COVID-19 pandemic, Chinese universities restrict students from going outside and limit their activities within campus. It thus causes students' complaints.}

\paragraph{Implicit Expressions.} The model struggles with complaints expressed in more subtle ways. These complaints do not contain any prominent negative words, but through other means like strike a chord or entrust. In the following example, the user hopes that managers can personally experience the status quo to understand the user's dissatisfaction. Therefore, predicting them correctly requires more contextual understanding: {\small 真的非常极其的希望校领导也能来感受一下封校的生活 (\textit{I really hope that school leaders can also come and experience the life of the closed school})}.

\paragraph{Vague Expressions.} We observe that the model is likely to be confused by vague or incomplete expressions in the posts. Consider the following example: {\small 赶上这破事，如果教育真的公平不如直接取消复试吧 (\textit{Encountered this shit. If the education is really fair, it is better to cancel the exam})}. The hypothetical relationship using {\small 如果 (\emph{if})} is an expression of uncertainty \cite{wei2018empirical} and there are no prominent emotional words in the text. Thus, our model fails to understand the speaker's intention well.

\section{Complaints as an Emotion}
\label{sec:complaint_senti}

From \Cref{tab:sample_tweets}, we notice stronger complaints seem to be associated with negative emotion words. Prior studies also point out that complaints can be treated as an influential emotional dimension \citep{iyiola2013relationship}. Then a natural question to ask is whether existing sentiment models have already been able to predict complaints intensity scores and our annotation efforts are actually not needed?

In this section, we demonstrate the necessity of building corpus annotated with complaints intensity, by showing the model trained on standard sentiment datasets fails to do well in our complaints intensity prediction task. We also show that analyzing complaints can be a useful complement for sentiment analysis.

\subsection{Differences between Complaints and Sentiment}
\label{sec:complaint_senti_diff}

In sentiment analysis, models normally output a score between 0 and 1, indicating how likely a post is to express negative emotion. Here we make the assumption that the most negative emotion may lead to the strongest complaint.
We first examine if these probability scores from sentiment models can be used as intensity scores for measuring complaints intensity.

\paragraph{Setup.} To ensure a fair comparison, we select a newly developed dataset on COVID-19, collected also from Weibo using hashtags related to COVID-19 by \citet{lyu2020sentiment}. The dataset contains 21,174 posts with fine-grained emotion annotations.\footnote{To the best of our knowledge, there does not exist a Chinese Weibo dataset annotated with continuous sentiment intensity. We note some datasets with discrete sentiment levels, like Douban movie short comments dataset \cite{ma2011recommender}.} 

In our experiments, we follow the same steps in \Cref{sec:collection} to pre-process this dataset. As our pre-processing steps result in a category imbalance issue, we merge categories with similar emotions. Specifically, we merge labels ``fear'', ``anger'', ``disgust'', and ``sadness'' into ``negative'' category and merge labels ``gratitude'', ``surprise'', and ``optimism'' into the ``positive'' category.
Finally, we have 8,783 posts in the negative category and 8,336 posts in the positive category.

We use BERT to train a sentiment model using the above COVID-19 data. We also sample 80\% of our corpus for developing our BERT-based complaints model. The performances of both models are compared on the left 20\% annotated posts. During evaluation, the sentiment scale from 0 to 1 is linearly mapped to our complaints intensity interval from -1 to 1.

\paragraph{Results.}  
Results are shown in \Cref{fig:sentiment model}. We observe that using the probability scores from sentiment models shows decent performance on our complaints intensity prediction task. It indicates a clear connection between complaints and emotions. We also observe that models trained on our annotated corpus outperform sentiment model, demonstrating the necessity of building such corpus for complaints intensity estimation. 

\begin{table}[h!]
\centering
\small
\begin{tabular}{lccc}
\toprule
\textbf{Model}          & \textbf{Preason's $r$}     & \textbf{MSE}     \\\midrule
Complaint & 0.76 & 0.20  \\
Sentiment & 0.71 & 0.24     \\
\bottomrule
\end{tabular}
\caption{Performances of sentiment model and complaint model for complaints intensity prediction task.}
\label{fig:sentiment model}
\end{table}

\paragraph{Valence and Arousal.}

We also quantitatively studied the correlation between complaints and sentiment through Valence-Arousal. Valence can be positive or negative and corresponds to the standard dimension of sentiment analysis; Arousal, which can be low or high and express the degree \cite{vorakitphan2020regrexit}. We use the VA score annotated by \citet{xu2021valence}, which contains 11,310 simplified Chinese words.
The valence and arousal ratings include scores -3 to +3 for valence rating and scores 0 to 4 for arousal rating. 

We identify sets of words in the Valence-Arousal lexicon that have high valence scores ($>$2), low valence scores ($<$-2), high arousal scores ($>$3), and low arousal scores ($<$2). A similar approach is used in \citet{hada2021ruddit}.
We average the scores of tokens from the above four dimensions in each post and calculate the correlation with our complaints intensity. 
Results in \Cref{tab:VA} show low valence and high arousal are more correlated with complaints intensity compared to other two dimensions.

\begin{table}[h!]
\centering
\small
\begin{tabular}{cc}
\toprule
\textbf{Dimension}    & \textbf{Pearson's $r$} \\\midrule
High valence & 0.02   \\
Low valence  & 0.31   \\
High arousal & 0.18   \\
Low arousal  & 0.05  \\
\bottomrule
\end{tabular}
\caption{Pearson's correlation between the complaints intensity scores and emotion dimensions.}
\label{tab:VA}
\end{table}

\subsection{Complaints Help Sentiment Analysis}
\label{sec:help_senti}

We now show that analyzing complaints could be helpful for the binary sentiment analysis task. 

\paragraph{Models.} 
We still use the COVID-19 dataset discussed in \Cref{sec:complaint_senti_diff} for the binary sentiment classification task.
We experiment with the SVM and BiLSTM-Attention models. The complaints score is added as an additional feature input to the model. 

\paragraph{Results.}

\Cref{tab:sentiment} shows the results of the models on the sentiment classification task. Overall, we observe that the models with the complaint feature perform better than the original model. It demonstrates that a simple add-on can boost the prediction accuracy of sentiment classification for non-neural and traditional neural models. 
We also provide the performance of BERT for reference in \Cref{tab:sentiment}.

\begin{table}[h!]
\centering
\small
\begin{tabular}{lccc}
\toprule
\textbf{Models}        & \textbf{P} & \textbf{R} & \textbf{F1} \\ \midrule
SVM                   & 0.51      & 0.49      & 0.50      \\
\, + complaint   & 0.53      & 0.50      & 0.51       \\\midrule
BiLSTM-Att                   & 0.72      & 0.70      & 0.71       \\
\, + complaint & 0.74      & 0.71      & 0.72      \\\midrule
BERT        & 0.79      & 0.76          & 0.77      \\
\bottomrule
\end{tabular}
\caption{Results for binary sentiment prediction. F1 score of models with complaint feature is significantly better than the original model ($p$-vlaue $<$ 0.01, $t$-test).}
\label{tab:sentiment}
\end{table}

\subsection{Case Study}

We are interested in what types of tokens sentiment model and complaint model try to capture. We thus take the BiLSTM-Attention model trained for sentiment classification task in \Cref{sec:help_senti} and our complaints model for comparison. We visualize the attention weights extracted from the above two models for the following example:
{\small 准备这么久的考试推迟真是绝了呵呵 (\emph{After preparing for so long, the exam is now postponed. It's absolutely speechless. Hmm. How interesting.})}.
We observe that sentiment model assigns high attention weights for tokens {\small 绝了 (\emph{speechless})} and {\small 呵呵 (\emph{Hmm. How interesting.})}, both expressing emotions. However, our complaint model puts high weights on tokens {\small 推迟 (\emph{postponed})} and {\small 考试 (\emph{exam})}. These are tokens that reflect the reasons for complaining. These differences again demonstrate the need for building a specific dataset for complaints intensity.

\begin{figure}[h!]
    \centering
    \includegraphics[width=0.47\textwidth]{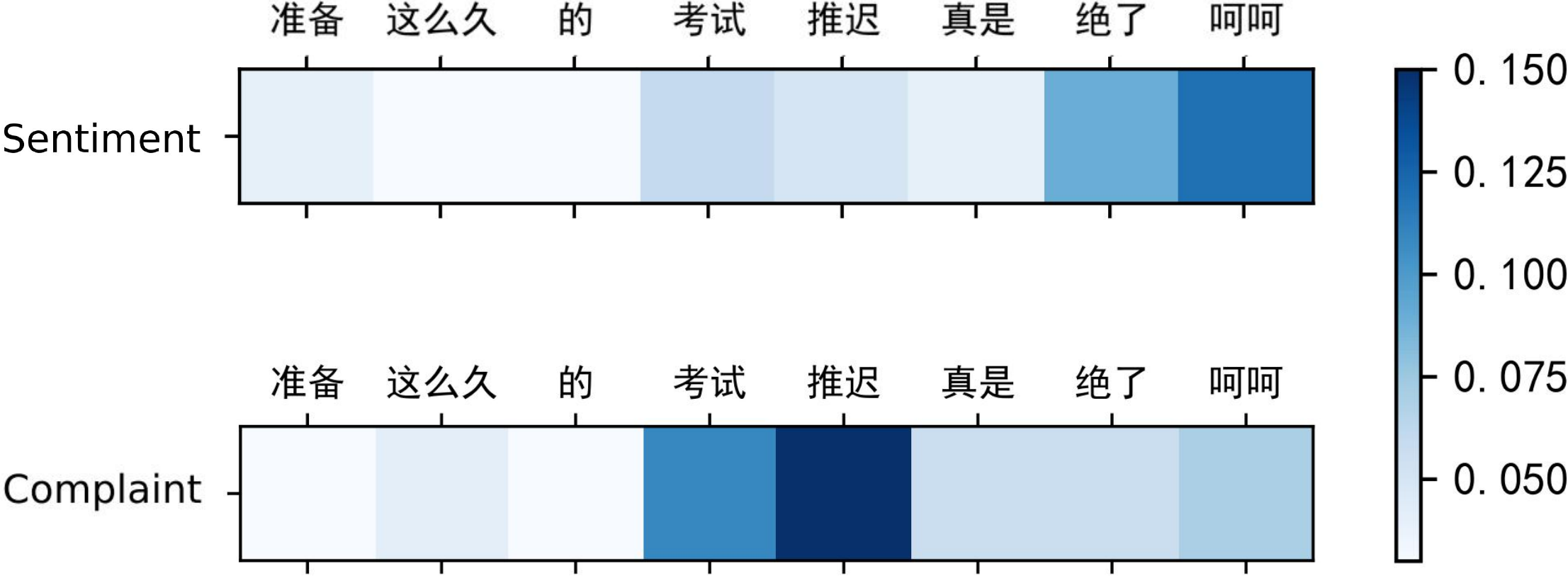}
    \caption{Attention weights for the sample sentences in sentiment model and our complaint model.} 
    \label{fig:attention}
\end{figure}

\section{Cross-lingual Analysis} 

Our newly collected complaints intensity dataset is written in Chinese, while current existing datasets \cite{preotiuc2019automatically, jin-aletras-2021-modeling} contain English tweets.
It provides us an opportunity to understand the linguistic differences for complaints made by Chinese and English speakers on social media. 

Our cross-lingual analysis is performed in the following way.
First, we evenly sample 200 complaint tweets released in \citet{jin-aletras-2021-modeling} from their defined four categories. Similarly, we divide our annotated data with intensity greater than 0 (as complaint posts) into 4 bins for sampling.
We ask 5 in-house annotators to mark and then use majority voting to decide if a post makes a direct or indirect complaint, along with the strategy used to make complaints. 

\paragraph{Direct and Indirect Complaints.}

According to \citet{boxer2010gripe}, the speech of complaint can be divided into direct and indirect complaints. In social media, direct complaints are addressed to a complainee who is held responsible for the complaint action, and the addressee is fully or partially responsible for this behavior. Here is an example for a direct complaint: {\small学校的管理怎么这么不人性化？ (\emph{Why is the management of the school so inhumane?})} 
While indirect complaints refer that the recipient is not primarily responsible for perceived complaints and is more about the evaluation of the target or event, such as: {\small这样做让我们的压力很大,真无语 (\emph{This puts us under a lot of pressure, so speechless})}.

We compare the percentages of direct and indirect complaints for two languages based on our annotations.
Results show that indirect expressions are more likely to be used in Chinese posts (80\% are indirect complaints). On the contrary, 91\% of English tweets make complaints in a direct way. This finding seems to be  consistent with the study of \citet{deng2019cultural}, which demonstrates that Chinese people tend to use indirect expressions.

\paragraph{Strategy.}
In pragmatics, strategy is an appropriate countermeasure adopted to achieve the purpose of language communication. \citet{1981Politeness} conducted a comparative study of English and German complaints from the direct degree and emotional markers. \citet{Anna2008} discussed the choice of Chinese and English complaining strategies and proposed an explicit-implicit strategy. Implicit complaints are very subtle and even use metaphors to express complaints about the target, which requires more semantic information to capture. In contrast, explicit complaints can be further divided into types of with-redress and without-redress. With-redress is the strategy of request for repair; without-redress usually contains complaint targets or objects, which can be easily identified by recognizing an obvious complaint word or phrase.

\Cref{tb:strategy} shows that strategy varies across languages. We find that the Chinese are more inclined to without-redress strategy, while the most frequent strategy used by Americans is with-redress strategies. It provides some empirical supports for findings in \citet{Anna2008}.

\begin{table}[h!]
\centering
\small
\begin{tabular}{lcc}
\toprule
\textbf{Strategy} & \textbf{Chinese} & \textbf{English}\\\midrule
Implicit  & 65\%             & 12\%   \\
With-redress    &   13\%   &   78\% \\
Without-redress &   22\%    &   10\%    \\
\bottomrule
\end{tabular}
\caption{Percentages of strategies across languages.}
\label{tb:strategy}
\end{table}

\paragraph{Irony.} Irony implies the opposite of the literal meaning. Dealing with non-literal means is a challenging task. In Twitter, \citet{reyes2012humor} used specific hashtags as gold labels to detect irony in a supervised learning setting, such as \textit{\#irony} and \textit{\#sarcasm}. \citet{attardo2013intentionality} observed that native speakers are usually able to process the meaning of sarcasm automatically, but the ability of second language learners to infer meaning from context varies greatly. In \Cref{sec:error_analysis}, we observe irony counts for the majority of errors made by our model.

We analyze the number of complaints using irony. To detect ironic expressions, we separately use the Chinese irony dataset of \citet{tang2014chinese} and the English dataset of \citet{van2018semeval} to train the Bi-LSTM model. Results showed that 10\% of Chinese data contained irony, and 26\% of English data contained irony. It shows that English speakers use ironic expressions more often compared to Chinese speakers. Further, we conduct part-of-speech analysis on these ironic expressions. \Cref{tab:pos} shows that Chinese irony has the highest proportion of nouns, followed by verbs; while in English irony, verbs are the most, followed by nouns. In addition, there are more adjectives and adverbs in English than in Chinese. 

\begin{table}[h!]
\small
\centering
\begin{tabular}{lcc}
\toprule
\textbf{Part of Speech} & \textbf{Chinese} & \textbf{English} \\\midrule
Nouns & 31.2\%   & 27.9\%             \\
Verbs & 21.8\%   & 35.2\%            \\
Adjectives & 3.1\%  & 10.7\%         \\
Adverbs & 9.9\% &  11.9\%           \\
\bottomrule
\end{tabular}
\caption{Percentages for POS tags across languages.}
\label{tab:pos}
\end{table}

\paragraph{Limitations.} We note a few limitations for our cross-lingual analysis. One limitation is domain mismatch. Our Chinese posts are collected from education domain, while English posts are collected from domains including food or online service. People may exhibit different behaviors when making complaints. We also note that the sample size for making comparisons is rather small, due to budget issues for experts annotations. In future work, we will perform a large-scale comparison by using the data collected from the same domains and utilizing crowdsourcing annotations.  

\section{Predicting Post Popularity}

Finally, we demonstrate that complaint intensity scores from our computational models can help estimate the post popularity on social media. We envision incorporating these scores into existing social media monitoring systems to improve their prediction accuracy.

\paragraph{Task.} Predicting the popularity of content on social media has been extensively studied in literature \cite{szabo2010predicting,hong2011predicting,bao2013popularity,carta2020popularity}. Our task is to predict the popularity of a Weibo post. Specifically, given the popularity prediction $p(t_{i-1})$ at time $t_{i-1}$, we wish to predict the popularity $p(t_i)$ at next time step $t_i$. The popularity $p(t)$ is measured by the number of blog posts under the topic at time $t$.

\paragraph{Methods.}
We follow \citet{szabo2010predicting} and consider the following baseline that only uses early prevalence for prediction:
\begin{equation}
\ln {\widehat p(t_{i})}= \alpha_1\ln p(t_{i-1})+\alpha_2,
\nonumber
\end{equation}
where $\alpha_1$ and $\alpha_{2}$ are learnable coefficients. It is justified as a strong baseline in \citet{bao2013popularity} that a strong correlation exists between logarithmically transformed popularity and early popularity.\footnote{\citet{bao2013popularity} further proposed to use link density and diffusion depth for popularity prediction. Above methods were tested on WISE 2012 challenge dataset. We tried our best but are not able to have access to this dataset. In our own collected dataset, we do not have link or diffusion information.}

The popularity of posts on social media can be measured by multiple dimensions \citep{bao2013popularity}. To show the effectiveness of our complaint scores, we add in the complaint intensity as a new term to estimate the final logarithmic popularity:
\begin{equation}
\ln {\widehat p(t_{i})}= \beta_1\ln p(t_{i-1}) + \underbrace{\beta_2 d_{c}(t_{i-1})}_{\text{complaints density}} + \beta_3,
\nonumber
\end{equation}
where $d_{c}(t_{i-1})$ is the complaints density at time $t_{i-1}$, calculated by the ratio between the sum of the complaint intensity of blog posts per unit time to the number of all blog posts. $\beta_1$, $\beta_2$ and $\beta_3$ are learnable new coefficients from data.

\paragraph{Setup.} We collected a new set of 4,973 posts under 8 hashtags on Weibo from March 2021 to November 2021 , which are shown in \Cref{tab:hashtags for application} (in \Cref{subsec:data for application}). We pre-process these posts using the same steps in \Cref{sec:collection}. Within each hashtag, 80\% of the posts are used for training, and the left 20\% are used for testing. We set the time step to be two hours. RMSE (root mean square error) and MAE (mean absolute error) are used to evaluate predicted results.

\paragraph{Results.}

We first examine the relationship between complaints density and post popularity as a sanity check. Results show a strong positive correlation 
with an upper cluster slope of 0.95. 

\begin{figure}[h!]
    \centering
    \includegraphics[width=0.46\textwidth]{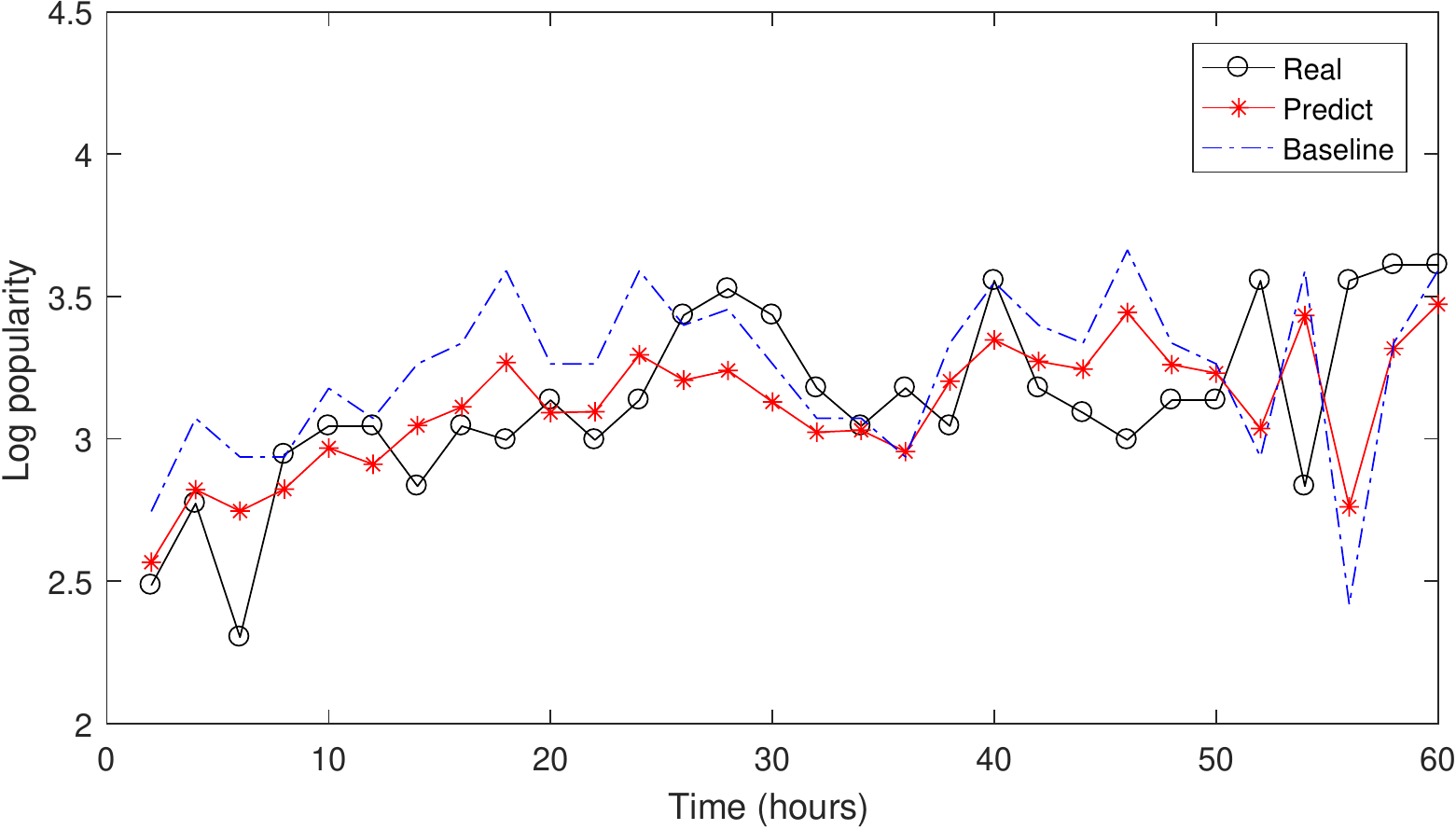}
    \caption{Comparison between actual post popularity and our predictions for hashtag {\small \emph{\#JuRen Education Group announces its bankruptcy\#}}. RMSE = 0.32, MAE = 0.30.}
    \label{fig:fit}
\end{figure}

We report our post popularity prediction results in \Cref{tab:application}. We observe our method that combines complaint density outperforms the baseline method. 
In \Cref{fig:fit} we also show the comparison between our predictions and real values for a specific hashtag: {\small \#巨人教育宣布倒闭\# (\emph{\#JuRen Education Group announces its bankruptcy\#})}.
We observe that adding complaints scores help better estimate the post popularity, especially in the early stages. 
It is probably because complaints are likely to draw users' attention to engage in discussions and hence boost the popularity of events.

\begin{table}[h!]
\small
\centering
\begin{tabular}{lcc}
\toprule
\textbf{Method}        & \textbf{RMSE} & \textbf{MAE} \\
\midrule
Baseline             & 0.41 $\pm$ 0.01 & 0.35 $\pm$ 0.01     \\
+ complaints density & 0.35 $\pm$  0.02 & 0.33 $\pm$ 0.01      \\
\bottomrule
\end{tabular}
\caption{RMSE and MAE for popularity prediction.}
\label{tab:application}
\end{table}

\section{Related Work}
\label{sec:related_work}

There have been various studies for complaints in linguistics, economics, and public opinion research. 
In \citet{olshtain1985complaints}, complaints are defined as what happened does not meet people's expectations, making people dissatisfied and blaming others.
\citet{kolodinsky1995usefulness} explained the characteristics of consumers' complaining behavior from an economic point of view.
\citet{liu2016optimizing} analyzed complaints about public transportation.
Complaints on social media have also drawn great attention in recent years.
\citet{andreassen2013online} focused on the study of the difference between social media complaints and traditional complaints, and argue that social media complaints are a unique way for consumers to express dissatisfaction.
\citet{balaji2015customer} studied the causes of complaints and found most of the complaints occur after a double deviation caused by dissatisfaction with the last solution. Motivated by these, in this paper, we collect complaints from Weibo, a widely used social media application.

In the area of linguistic studies on computational sociology, \citet{Meinl2010Electronic} studied the complaints act sequence in eBay reviews through 200 annotated English and German reviews. \citet{ganesan2016linguistic} collected 2,500 reviews from Yelp and Walmart about commodity, then manually categorized it into 5 categories: negative only, complaint, positive only, raise, and irrelevant. 
\citet{preotiuc2019automatically} focused on binary classification between complaints and non-complaints in various domains, such as food, car, online service, e-commerce.
\citet{jin-aletras-2021-modeling} categorized complaints into 4 categories: no explicit reproach, disapproval, accusation, and blame.
In this work, we present the first study of estimating the intensity of complaints from text.

Our work is also related to prior work on emotion detection and sentiment intensity estimation \citep{mohammad-bravo-marquez-2017-wassa, cortis-etal-2017-semeval}.
\citet{kiritchenko-mohammad-2017-best} created a Twitter dataset annotated with sentiment intensity. 
We have discussed the connections between complaints and sentiment in detail in \Cref{sec:complaint_senti}.

\section{Conclusion}
\label{sec:conclusion}

In this paper, we present the first study of measuring the intensity of complaints from text. 
We build a corpus of 3,103 Chinese Weibo posts about complaints,
annotated with complaints intensity scores using Best-Worst Scaling method. 
We then demonstrate that our corpus supports the development of automatic computational models for accurate complaints intensity predictions. 
Furthermore, we study the connections between complaints and sentiment, and perform a cross-lingual comparison for complaints expressions between Chinese and English. 
We finally show that our complaints intensity scores help better estimate the posts popularity on social media.

\section*{Ethical Concerns}

In this paper, we collect a complaint dataset from Weibo. The tools we use to collect posts comply with Weibo's terms of service. 
We will follow Weibo's policy for content redistribution to release our annotated corpus.
Specifically, we will not release any user information or demographic data, including the authors' names, ages, and origins.

We recruited part-time research assistants for our annotation task. Annotators were warned that the complaint posts might contain offensive or upsetting content. Annotators were shown only anonymized posts and agreed not to make attempts to de-anonymize them. We did not collect any personal data from the annotators before, after, or during the annotation task. Moreover, we pay them 15.7 USD/hour and at most 14 hours per week. 

\section*{Acknowledgments}

We would like to thank the anonymous reviewers for their valuable feedback. 
This work was partially supported by a grant from the Research Grants Council of the Hong Kong Special Administrative Region, China (Project No. PolyU/25200821) and CCF-Baidu Open Fund (No. 2021PP15002000).

\bibliography{anthology}
\bibliographystyle{acl_natbib}

\onecolumn

\setcounter{table}{0}
\renewcommand{\thetable}{A\arabic{table}}

\appendix

\section{More Sample Posts}
\label{subsec:keywords}

We provide more sample posts in \Cref{tab:sample_more} from our dataset grouped according to the 5 bins defined in \Cref{tab:sample_tweets}.

\begin{table*}[h!]
\centering
\resizebox{0.99\textwidth}{!}{
\begin{tabular}{cp{0.92\textwidth}c}
\toprule
\textbf{Bin} & \textbf{Weibo posts}  & \textbf{Scores}\\\midrule
1 & {\small 突然感觉农业大学做的还不错 (\emph{Suddenly I feel that the Agricultural University is doing pretty well.})}     &  {\small -1.00}                                              \\
1 & {\small 校内完成作业挺好的，回家可以做自己喜欢的事  (\emph{It’s good to finish homework in school, you can do what you like when you go home.})}     &  {\small -0.80}                   \\\midrule
2   & {\small 为了疫情防控而封闭管理其实是没有问题的。但是既然实行这项制度，就要真正把好关，校外人员不能随意进出学校 (\emph{In fact, there is no problem with closed management for the prevention and control of the epidemic. However, since this system is implemented, it is necessary to truly ensure that people outside the school cannot enter and leave the school at will.})}           & {\small-0.50}       \\
2   & {\small 让学生在校内写完作业这不怎么可能实现吧？对家长来说确实挺好的，因为很多题都不会，也没办法辅导。 (\emph{Isn’t it possible for students to finish their homework in school? It's really good for parents, because many problems parents don't know how to solve them, and they can't help students.})}           & {\small-0.33}       \\\midrule
3   & {\small 从来没有人去教家长要如何做一个合格的学生家长 (\emph{No one has ever taught parents how to be a qualified student parent.)}}           & {\small-0.15}       \\
3   & {\small 这不太好吧，所谓家庭作业不是在家完成么，在校内完成的不是校内作业或者课堂作业吗，真是这样那干脆不要布置家庭作业罢了 (\emph{This is not so good. Isn't the so-called homework done at home? Isn't it done in school or classwork? If it's true, then just don't assign homework.})}           & {\small+0.17}       \\\midrule
4   & {\small 老师也是人他们虽然是服务行业但也需要自己的生活吧小学作业也没有那么多根本不用老师熬夜去批那初中毕业学年和高中呢作业量多难度大一个老师交两三个班一百多号人学生写完作业都要十点多了难道还让老师在学校通宵批完没有效率出现错误又说是老师不负责?  (\emph{Teachers are also humans. Although they are in the service industry, they also need their own lives. There are not so many primary school homework. There is no need for teachers to stay up late to correct them. The middle and high school homework is a lot and difficult. One teacher teaches two or three classes. For many students, it’s more than ten o’clock when the students finish their homework. Could it be that the teacher is allowed to finish the correction at school overnight? This is not efficient. If there is a mistake, the teacher will be accused of being irresponsible.})} &   {\small+0.40}   \\
4   & {\small 为什么都是0分啊？是作弊被抓到了吗？还是怎么样？还是根本就没来考试啊？浪费这机会，有那机会给我多好，我想上还上不了呢 (\emph{Why are their scores all 0 points? Was it caught for cheating? Or what? Or didn’t you come to the exam at all? The two of them wasted this opportunity, how good it is for me to have that opportunity, I want to go to school but I don’t have the opportunity.})} & {\small+0.57}   \\
\midrule
5 & {\small \textbf{学校}偏僻，所以西安这些学校的职工都是从隔壁村子随便找的？隔壁西电，\textbf{门卫}满嘴官话实际怠惰工作，\textbf{保洁}在图书馆大声唠嗑，合着招职工没有限制应聘即上岗?西外\textbf{职工}都敢拖行女生了，原来职工素质低不是我校特例啊西安高校，你有事吗? (\emph{Due to \textbf{the school}'s remoteness, the employees of these schools in Xi'an are all looking for them from the neighboring village? At the Xidian University next door, \textbf{the guards} are lazy, and \textbf{the cleaners} babble loudly in the library. Is it possible to recruit staff to take up jobs without restrictions? The school's \textbf{security guards} are very rude to girls. It turns out that the low quality of staff is not a special case of our school. The problems in Xi'an colleges and universities are severe.})}      & {\small+0.92}  \\
5   & {\small 现在的学校，真恶心，老师们拿的有工资啊，双休寒暑假，还美其名曰：家校共育这本来没错，但，能不能不要让家长充当老师的角色？？？？作业，回家写可以，家长还要拍照、打卡、发视频交作业，还要批改作业，这都是老师的工作好吗？？这部分工作家长做了，老师在干啥？这部分工作的工资，学校给家长发了吗？？没有啊！！那就请老师们，完成你们份内的工作！！别说什么辛苦之类的，拿着那份工资与待遇，就要干好那份工作在其位不谋其职！！！！ (\emph{The current school is really disgusting. The teachers are paid, and they have two winter vacations and summer vacations. They also have a good name: family-school co-education is not wrong. But can we not let parents act as teachers? ? ? ? Homework can be written at home. Parents have to take photos, check in, send videos to hand in homework, and also correct homework. This is the teacher's job. If the parents do this part of the work, then what does the teacher do? Has the school sent the salary for this part of the work to the parents? ? No! ! Then please teachers, finish your job! ! Not to mention that the work is very hard. With the salary and benefits, you must do the job well. Don't be irresponsible in this position! ! ! !})} & {\small+1.00}    \\
\bottomrule
\end{tabular}
}
\caption{More sample posts for each of the 5 bins. Words in bold are some points of concern.}
\label{tab:sample_more}
\end{table*}

\clearpage

\section{Data Used for Post Popularity Prediction}
\label{subsec:data for application}
We collected blog posts under 8 topics from Weibo to verify the relationship between complaint density and popularity. 
\Cref{tab:hashtags for application} shows the hashtag contents, along with the number and time of collection.

\begin{table}[h!]
\resizebox{0.99\textwidth}{!}{
\begin{tabular}{lcl}
\toprule
\multicolumn{1}{c}{\textbf{Hashtag}}                                         & \textbf{Number} & \textbf{Start From} \\
\midrule
\#巨人教育宣布倒闭\# (\emph{\#JuRen Education announces bankruptcy\#})                              & 1,663            & 2021/8/31           \\\midrule
\begin{tabular}[c]{@{}l@{}}\#官方回应开学典礼学生晕倒无人扶\#\\ (\emph{\#The official responded to the situation where the student fainted and no one helped at the opening ceremony\#})\end{tabular} &
  331 &
  2021/9/2 \\\midrule
\begin{tabular}[c]{@{}l@{}}\#芝加哥大学24岁中国留学生被枪杀\#\\ (\emph{\#A 24-year-old Chinese student at the University of Chicago was shot dead\#})\end{tabular} &
  179 &
  2021/11/11 \\\midrule
\begin{tabular}[c]{@{}l@{}}\#教育部将抑郁症筛查纳入学生体检\#\\ (\emph{\#Ministry of Education incorporates depression screening into student physical examination\#})\end{tabular} &
  1,023 &
  2021/10/31 \\\midrule
\#江苏一建停考\# (\emph{\#Jiangsu Province Level One Architect Examination Suspended\#})         & 286             & 2021/8/24           \\\midrule
\#计算机二级证书有必要吗\# (\emph{\#Is it necessary to take the second-level computer certificate\#}) & 242             & 2021/11/9           \\\midrule
\#西安外国语大学封闭管理\# (\emph{\#Closed management of Xi’an International Studies University\#})   & 287             & 2020/9/19           \\\midrule
\#大连理工大学支教\# (\emph{\#Supporting Teaching at Dalian University of Technology\#})            & 962             & 2021/3/8           \\
\bottomrule
\end{tabular}
}
\caption{The hashtag and its number used in the application.}
\label{tab:hashtags for application}
\end{table}

\end{CJK}
\end{document}